\documentclass{article} %
\usepackage[nonatbib,preprint]{neurips_2019}

\usepackage[T1]{fontenc}    %
\usepackage{url}            %
\usepackage{booktabs}       %
\usepackage{amsfonts}       %
\usepackage{nicefrac}       %
\usepackage{microtype}      %

\usepackage{soul}
\usepackage{common}
\usepackage{tikz}
\usepackage{xspace}

\usepackage{hyperref}       %
\usepackage{xcolor}
\hypersetup{
    colorlinks,
    linkcolor={red!0!black},
    citecolor={blue!80!black},
}
\usepackage[nameinlink]{cleveref}
\usepackage[numbers]{natbib}
\usepackage{subcaption}

\DeclareMathOperator{\tr}{tr}

\newcommand{\E}{\mathbb{E}}
\newcommand{\w}{w}

\renewcommand{\H}{\mathcal{H}}
\newcommand{\D}{\mathcal{D}}
\renewcommand{\L}{\mathcal{L}}
\newcommand{\R}{\mathbb{R}}
\renewcommand{\div}{\operatorname{div}}
\newcommand{\Y}{\mathcal{Y}}
\newcommand{\X}{\mathcal{X}}

\newcommand{\set}[1]{\left\{#1\right\}}

\newcommand{\bra}[1]{\left[#1\right]}

\newcommand{\norm}[1]{\left\|#1\right\|}

\renewcommand{\v}{\mathbf{v}}

\newcommand{\cut}[1]{{}}

\newcommand{\utext}[2]{\underbrace{#1}_{\text{#2}}}
\newcommand{\KL}[2]{\operatorname{KL}(\,#1\,\|\,#2\,)}

\newcommand{\eff}{{\text{\textit{eff}}}}

\def\eg{\textit{e.g.}\@\xspace}
\def\ie{\textit{i.e.}\@\xspace}

\newcommand{\be}{\begin{equation}}
\newcommand{\ee}{\end{equation}}

\title{
Dynamics and Reachability of Learning Tasks
}

\author{Alessandro Achille${}^*$ \quad Glen Bigan Mbeng${}^\dag$ \quad Stefano Soatto${}^*$\\ *University of California, Los Angeles; ~~~~~~ $\dag$ SISSA and INFN\\
\texttt{achille@cs.ucla.edu, gmbeng@sissa.it, soatto@cs.ucla.edu}
}

\begin{document}

\maketitle
\begin{abstract}
  We compute the transition probability between two learning tasks, and show that it decomposes into two factors. The first depends on the geometry of the loss landscape of a model trained on each task, independent of any particular model used. This is related to an information theoretic distance function, but is insufficient to predict success in transfer learning, as nearby tasks can be unreachable via fine-tuning. The second factor depends on the ease of traversing the path between two tasks.  With this dynamic component, we derive strict lower bounds on the complexity necessary to learn a task starting from the solution to another, which is one of the most common forms of transfer learning.
\end{abstract}

\section{Introduction and related work}

Among the many virtues of deep neural networks is their {\em transferability}: One can train a model for a task ({\em e.g.,} finding cats and dogs in images), and then use it for another ({\em e.g.,} outlining tumors in mammograms) with relatively little effort. Sometimes it works. Alas, little is known on how to predict whether or not such \emph{transfer learning} or \emph{domain adaptation} will work, and if so how much effort is going to be needed, without just trying-and-seeing -- a process that has been referred to as ``alchemy''. It is not a given that training on a sufficiently rich task, and then fine-tuning on anything else, must succeed. Indeed, slight changes in the statistics of the data can make the optimal solution to a task {\em unreachable}  \cite{achille2018critical}.%
\footnote{We introduce the notion of \emph{reachability} of a task in \Cref{sec:dynamic-distance}.}

At the most fundamental level, understanding transfer learning or domain adaptation requires understanding the \emph{topology and geometry of the space of tasks.} When are two tasks ``close''? Can one measure the distance between tasks without actually running an experiment? Does knowing this distance help predict whether transfer learning is possible, and if so how many resources will be needed?

This has motivated recent interest in defining distances between classification tasks, but there are shortcomings. Architecture independent distances, such as lexicographic distances between label sets in a taxonomy, fail to capture the complex learning dynamics of deep neural networks (DNNs), which can fail in adapting to slight perturbations of the data distributions, even if the task variable remains identical \cite{achille2018critical}. On the other hand, distances between parametric representations of a task, for instance the weights of DNNs trained on them, fail to capture that very different parameters can represent the exact same posterior distribution. In order to relate to transfer learning, a distance function would have to be asymmetric \cite{mennucci2007asymmetric} as it is typically easier to fine-tune a simple task from a complex one than vice-versa.  Such distances can be defined \textit{a posteriori} by looking at the performance of fine-tuning \cite{zamir2018taskonomy}, which however fails to highlight whether the datasets themselves have a distance structure, and how this interacts with the training process of the deep network.

In this paper, we tackle both questions. We use the results of \cite{achille2019information}, which introduces an intrinsic asymmetric distance on the space of learning tasks based on the amount of structure one needs to learn, and which is independent of the particular learning algorithm. While such a ``static distance'' gives qualitatively good results in many cases \cite{achille2019task2vec}, it does not fully capture problems, particularly of domain adaption, where even nearby tasks may be \emph{unreachable} with fine-tuning \cite{achille2018critical}.  We then show how the \emph{dynamics} of the learning process, that is, of SGD, interact with the topology of the space of learning tasks, and in particular how the distance between learning tasks relates to transfer learning.

We therefore characterize the probability and expected training time of reaching one task from another, using mathematical tools from physics, in particular Kramer's rate theory \cite{kramers1940brownian, caroli1981diffusion} and the path-integral approach \cite{hunt1981path}, which allows us to model the probability of different paths that SGD may easily take to reach a a solution to the task. We then show that, to first approximation, the transition probability from the solution to a first task (pre-training) to the solution of a second task (fine-tuning) factorizes into two parts. Surprisingly, one turns out to be precisely the intrinsic ``static'' distance on the space of tasks. The other, which we call {\em dynamic} distance, depends on the existence in the loss-landscape of likely path that the optimization procedure (SGD) can use to  {\em reach} a task from another. It should be noted that, while the asymptotic dynamics of SGD are well studied \cite{chaudhari2018stochastic}, our focus is on the initial convergence phase, that has received relatively little attention in the literature thus far.

Finally, we verify empirically that the distance we define correlates with the ease of transfer learning and the convergence time on a task.

\section{Preliminaries and notation}

In supervised learning, we are given a training dataset $\D = \set{(x_i,y_i)}_{i=1,\ldots,N}$ of $N$ samples, where  $x_i \in \X$ is the observed input data ({\em e.g.}, an image) and $y_i \in \Y$ is an hidden random variable, our \emph{task} that we are training to reconstruct ({\em e.g.}, a label).

A Deep Neural Network trained is a  family of functions, parametrized by \emph{weights} $w$, that encodes a posterior probability $p_w(y|x)$ of the task variable $y$ given the input.
The weights $w$ are usually optimized to minimize the cross-entropy loss $L_\D(w) = \E_{(x,y) \sim \D}[-\log p_w(y|x)]$ on the training set $\D$.

Exploiting the additive structure of $L_\D(w)$, the cross-entropy loss is usually minimized using \textit{stochastic gradient descent} (SGD), which, rather than performing gradient descent using the full gradient $\nabla_w L_\D(w)$, updates the weights $w$ with a cheaper gradient estimate computed from a small number of samples (mini-batch). That is, $w_{k+1} = w_k - \eta \nabla \hat{L}_{\xi_k}(w)$,
where $\xi_k$ are the indices of a randomly sampled mini-batch of size $|\xi_k| = B$, and
$\hat{L}_{\xi_k}(w) = \frac{1}{|\xi_k|} \sum_{i \in \xi_k} [-\log p_w(y_i|x_i)]$.

Notice that the mini-batch gradient $\nabla \hat{L}_{\xi_k}(w)$ is an unbiased estimate of the real gradient, \ie, $\E_{\xi_k}[\nabla \hat{L}_{\xi_k}(w)] = \nabla L(w)$. Hence, we can think of $\nabla L_{\xi_k}(w)$ as a noisy version of the real gradient.
This allows us to rewrite the update equation of SGD as
\begin{equation}
\label{eq:sgd-diffusion}
w_{k+1} = w_k - \eta \nabla \hat{L}_{\xi_k}(w_k) + \sqrt{\eta}\, T_{\xi_k}(w_k),
\end{equation}
where we have introduced the noise term $T_{\epsilon_k}(w) = \sqrt{\eta}\,\big(\nabla \hat{L}_{\xi_k}(w) - \nabla L(w)\big)$. Written in this form, \cref{eq:sgd-diffusion} is a Langevin diffusion process \cite{li2015stochastic,chaudhari2018stochastic}, a fact that plays a central role in our analysis. While it is known that the noise term $T_{\xi_k}(w_k)$ is non-Gaussian and non-isotropic for standard DNNs \cite{chaudhari2018stochastic}, modeling $T_{\xi_k}(w_k)$ as uncorrelated white noise still provides useful intuitions in the analysis of SGD while simplifying the theoretical analysis.
In the limit of small step-size, \cref{eq:sgd-diffusion} can be approximated as the continuous stochastic process \cite{li2015stochastic}:
\[\dot{\w} = f(\w) + \sqrt{2 D} n(t),\]
where $D \propto \eta/B$ is a dissipation constant. We use this approximation in throughout the paper.

We also make use of the  Kullbach-Liebler (\textit{KL}) divergence $\KL{p(x)}{q(x)}$ between  distributions $p(x)$ and $q(x)$, which is defined as $\KL{p(x)}{q(x)} := \E_{x\sim p(x)}\big[\log (p(x)/q(x))\big]$. We recall that the KL-divergence is always non-negative, and it is zero if and only if $p(x) = q(x)$ \cite{cover2012elements}. Intuitively, it  measures the (asymmetric) similarity between two distributions. Given a family of conditional distributions $p_w(y|x)$ parametrized by a vector $w$, we can ask how much a small perturbation $\delta w$ of the parameters $w$ will change the distribution. To second-order, the divergence between the original and perturbed distribution is given by
\[F:=\E_x\KL{p_w(y|x)}{q_{w+\delta w}(y|x)} = \delta w^t F \delta w + o(\|\delta w\|^2)\]
where $F$ is the \textit{Fisher Information Matrix}, defined as
\[F = \E_{x, y\sim p(x)p_w(y|x)}[\nabla \log p_w(y|x)^t\nabla \log p_w(y|x)] = \E_{x\sim p(x)p_w(y|x)}[- \nabla^2_w \log p_w(y|x)].\]
For its relevant properties see, \eg, \cite{martens2014new}. Notice that the Fisher depends on the ground-truth data distribution $p(x,y)$ only through the domain variable $x$, not the \emph{task variable} $y$, since $y\sim p_w(y|x)$ is sampled from the model distribution when computing the Fisher.

\section{The Structure Function of a Task}

We consider a learning task to be implicitly defined by the training dataset $\D = \{ x_i, y_i\}_{i=1}^N$ and loss function that is provided to us. A natural question is when two different tasks $\D$ and $\D'$ are close to each other. Note that two datasets may not share any sample, yet define the same task. Moreover, most of the information contained in the dataset will usually be about nuisances (such as the foliage of the trees, objects in the background), which are not relevant for the task, which furthers complicates defining a distance on the information that actually matters.

To address these problems, \cite{achille2019information}, based on previous work by \cite{vereshchagin2004kolmogorov} on algorithmic information theory,  introduces a notion of \emph{structure} of a task, which serves to separate nuisance variability from the task-relevant information. They then define the (asymmetric) distance between two tasks as the amount of additional structure one needs to learn in order to solve the second task given a solution to the first.
Formally, for a given prior $P(w)$ the Structure Function of a dataset is defined by:
\begin{equation}
\label{eq:structure-function}
S_{\D}(t) = \min_{\KL{Q(w|\D)}{P(w)} < t} \E_{w \sim Q(w|\D)}[L_\D(w)].
\end{equation}
where the minimization is done over a ``posterior'' distribution $Q(w|\D)$ over the weights, which can depend on the dataset $\D$. Intuitively, the structure function express the optimal trade-off between information stored in the parameter of the model and error in the task. Since precisely codifying a parameter vector $w$ requires infinite information, we allow a ``noisy'' parameter distribution $Q(w|\D)$. Then,  $\KL{Q(w|\D)}{P(w)}$ represent the amount of bits to encode  $Q(w|\D)$ relative to the prior.

As we increase the amount of information that we store in the model, we can fit the dataset progressively better and $S_\D(t) \to 0$. However, this is subject to a diminishing return where more and more information needs to be encoded, in order to reduce the loss. To study this trade-off, it is useful to introduce the  Lagrangian corresponding to the minimization problem of \cref{eq:structure-function}:
\begin{equation}
\label{eq:kl-complexity}
C_\beta(\D; P, Q) = \E_{w \sim Q(w|\D)}[L_\D(p_w(y|x))] + \beta  \KL{Q(w|\D)}{P(w)}.
\end{equation}
\cite{achille2019information} shows that there is a critical value of the Lagrange multiplier $\beta$, such that the model stops encoding structural information about the task, that is, information features that can generalize, and starts memorizing nuisances of the training set. This suggests defining the Information in the Weights that minimizes $C_\beta(\D; P, Q)$ for $\beta$ as the amount of structural information of the task at level $\beta=1$.
Notice that this quantity is closely related to the Information Bottleneck of the Weights studied in \cite{achille2017emergence}, to PAC-Bayes theory \cite[Theorem 2]{mcallester2013pac}. In particular, for $\beta=1$, it reduces to the Evidence Lower Bound (ELBO) of Variational Inference.

\section{Static distance between tasks}
\label{sec:reachability}

Given two datasets, $\D_1$ and $\D_2$, we may consider the dataset $\D_1 \cup \D_2$ obtained by concatenating them. The amount of extra information that we need to learn $\D_2$ after learning $\D_1$ is the difference in structure between $\D_1 \cup \D_2$ and $\D_1$, suggesting the following definition of distance \cite{achille2019information}
\[
d_\beta(\D_1 \to \D_2) = C_\beta(\D_1 \cup \D_2; P) - C_\beta(\D_1; P),
\]
where $C_\beta(\D; P) = \min_{Q(w|\D)} C_\beta(\D; P, Q)$.

Depending on the choice of the prior $P(w)$, we can obtain different instantiations of $C_\beta(\D; P, Q)$ and hence of the distance $d_\beta(\D_1 \to \D_2)$. A particularly appealing choice, both for its simplicity and its connections to SGD dynamics as we will show later, is to pick a Gaussian prior $P(w) = N(0, \lambda^2 I)$ and a Gaussian posterior $Q(w|\D)=N(w_0, \Sigma)$, in which case we have the closed-form expression:
\[
\KL{Q(w|\D)}{P(w)} = \frac{1}{2} \bra{\frac{w_0^2}{\lambda^2} + \frac{1}{\lambda^2} \tr{\Sigma} + k \log{\lambda^2} + \log(|\Sigma|) - k},
\]
We now are interested in finding the distribution $Q(w|\D)$ that minimizes $C_\beta(\D; P, Q)$. Of course, finding the optimal weights $w_0$ is far from trivial, as it involves training a deep network on the dataset. However, we can give a description of the optimal $\Sigma$ for a weight configuration $w_0$: Approximating $C_\beta(\D; P, Q)$ to the second order at $w_0$ and minimizing in $\Sigma$, we obtain the minimizer:
\[\Sigma^*=\frac{\beta}{2} (H + \frac{\beta}{2\lambda^2} I)^{-1},\]
where $H$ is the Hessian of $L_\D(w)$. Using this, we obtain the following expression for $C_\beta(\D; P, Q)$ as a function of the local minimum $w_0$:
\begin{equation}
\label{eq:approximated-beta-complexity}
C_\beta(w_0) = C_\beta (\D; Q, P) = L_\D(w_0) + \frac{\beta}{2} \bra{\frac{\norm{w_0}^2}{\lambda^2} + \log \left|\frac{2\lambda^2}{\beta} H +  I \right|},
\end{equation}
where $H$ is  computed in $w_0$. As this approximation requires $H$ to be positive semi-definite, which it may not always be, we follow \cite{martens2014new} and rather use the Fisher Information Matrix as a robust positive semi-definite approximation of $H$. Note that this links the ``information complexity'' of the task to the local curvature of the loss landscape at that point. that is, its Hessian or its Fisher Information.

\section{The Dynamic Distance between tasks}
\label{sec:dynamic-distance}

In the previous section, we defined a notion of a ``static'' distance between tasks that is independent of the optimization algorithm used.
But how difficult is it for a SGD to find a solution to task $\D_2$ starting from task $\D_1$? That is, how difficult is it to fine-tune? In this section, we approximate the dynamics of SGD to quantify the extent in which the static distance and the learning dynamics affect the ease of fine-tuning.

Consider a network trained with the $L_2$ regularized loss $U(\w) = L_\D(w) + \gamma/2 \norm{w}^2$.
By taking the continuous limit of \cref{eq:sgd-diffusion}, that is, by letting the step size go to zero, we obtain that sample paths evolve according to the stochastic differential equation (SDE) \cite{li2015stochastic}
\[\dot{\w} = f(\w) + \sqrt{2 D} n(t),\]
where $f(\w) = \nabla U(w)$, $D$ is a constant and $n$ is the derivative of a Wiener process.
Given the SDE, we can derive a probability functional over paths, using the Martin-Siggia-Rose formalism. More precisely, the probability of a path $w(t): \R \to \mathcal{W}$ starting from $w_0$ at time $t_0$ is given by \cite{caroli1981diffusion}:
\begin{equation}
\label{eq:msr}
p(\w(t)|\w_0, t_0) = e^{-S(\w(t))} = e^{-\int_{t_0}^{t_f} \L(\w(t), \dot{\w}(t)) dt},
\end{equation}
where we have defined the Onsager-Machlup Lagrangian
\begin{equation}
\L(\w(t), \dot{\w}(t)) = \frac{1}{4D} \norm{\dot{\w}(t) - f(\w)}_2^2 + \frac{1}{2}\div f(\w).
\end{equation}
Notice that the density function in \cref{eq:msr} penalizes paths whose speed does not match the gradient $f(w)$, and adds a correction based on the divergence of the gradient field in order to account for concentrating or dissipating effects of the potential, which relates to the curvature of the energy $U$.

One of the main objects of interest for us is the transition probability $p(\w_f,t_f|\w_0, t_0)$ between two points $w_0$ and $w_f$ in time $\Delta t = t_f - t_0$. This can be expressed, given the probability distribution over paths, as
\begin{equation}
\label{eq:transition}
p(\w_f,t_f|\w_0, t_0) = \int_{\w_0}^{\w_f} p(w(t)|w_0, t_0) d\w(t) = \int_{\w_0}^{\w_f} e^{-\int_{t_0}^{t_f} \L(\w(t), \dot{\w}(t)) dt} d \w(t),
\end{equation}
where the integral is over all paths $w(t)$ such that $w(t_0)=w_0$ and $w(t_f) = w_f$. That is, the probability of reaching $w_f$ at the given time is the mass or ``volume'' of all paths reaching $w_f$. Estimating this, gives us information on which part of the loss landscape are easily {\em accessible}, or reachable, via SGD in a given training time.

Intuitively, we may expect the probability of reaching a point to depend on two separate factors: The energy gap between the initial and final configurations, as well as the existence of probable paths connecting them. To see this formally,  notice that the path density \cref{eq:msr} can be rewritten using the Stratonovic convention \cite{karatzas2012brownian} as
\begin{align*}
p(\w(t)|\w_0, t_0)
&= e^{-\int_{t_0}^{t_f} \L(\w(t), \dot{\w}(t)) dt} \\
&= e^{-\int_{t_0}^{t_f} \frac{1}{4D} \norm{\dot{\w}(t) - f(\w)}_2^2 + \frac{1}{2} \div f(\w) dt} \\
&= e^{-\int_{t_0}^{t_f} \frac{1}{4D} [\dot{\w}(t)^2 -2 f(\w)\dot{\w}(t) + f(\w)^2] + \frac{1}{2} \div f(\w) dt}\\
&= e^{-\frac{1}{2D} [U(\w(t_f)) - U(\w(t_0)] } e^{-\frac{1}{2D}\int_{t_0}^{t_f} \frac{1}{2} [\dot{\w}(t)^2 + f(\w)^2] + D \div f(\w) dt},
\end{align*}
We define the \emph{effective potential} $V(w)$ as
\begin{equation}
V(w) = \frac{1}{2} f(w)^2 + D \div f(\w) = \frac{1}{2} \nabla U(w)^2 - D \nabla^2 U(\w),
\end{equation}
so we can write
\begin{equation}
p(\w(t)|\w_0, t_0) = e^{-\frac{1}{2D}
\Delta U} e^{-\frac{1}{2D}\int_{t_0}^{t_f} \frac{1}{2} \dot{\w}(t)^2 + V(\w(t)) dt}.
\end{equation}

\subsection{Reachability of a task}

Substituting this expression in \cref{eq:transition}, we obtain a corresponding decomposition for the transition probability
\begin{equation}
\label{eq:decomposition}
p(\w_f,t_f|\w_0, t_0) = \utext{e^{-\frac{1}{2D}
[U(w_f) - U(w_0)]} }{Static potential}   \int_{\w_0}^{\w_f} \utext{ e^{-\frac{1}{2D}\int_{t_0}^{t_f} \frac{1}{2} \dot{\w}(t)^2 + V(\w(t)) dt}}{Reachability} d\w(t).
\end{equation}
The first part is \emph{static} in the sense that it depends only on the initial and final configurations and is independent of the path used to reach it.
The second factor measures existence of likely paths $\w(t)$ connecting the two points. It is called reachability because, regardless of how large the drop in static potential, the absence of probable paths makes transfer learning unlikely to succeed.

\subsection{Curvature, most likely paths and Lagrange approximation}

In principle, reachability depends both on the task ({\em i.e.,} the data), and the architecture. However, we will now show that to first-order approximation it depends only on information theoretic quantities. In particular, we show that, to first approximation, the most likely path is deterministic  and follows an effective potential $U_\eff(w) = U(w) - D \log |\H U(w)|_+$, where $|\H U(w)|_+$ denotes the determinant of the  positive part of the Hessian, \textit{i.e.}, the product of all positive eigenvalues. That is, {\em the potential needs to be corrected in order to account for the local curvature and the amount of correction depends on the temperature. }

One consequence of this fact is that sharp minima may not be minima at all for this particular potential when the temperature is sufficiently high. We will also show that the dynamic part of the potential can create spurious local minima that can inhibit learning of new problems in a transfer learning scenario. We will later also connect the curvature to the amount of information needed to solve a task. Using this connection, we will be able to characterize the {\em ``learnability''} (reachability)  of a task in terms of information-theoretic properties of the data. This completes our program of characterizing the geometry and topology of the space of tasks in a manner that does not depend on how the task is actually learned.

To start, we make a Lagrange (or Saddle Point) approximation: given two points $w_0$ and $w_f$, we assume that the probability concentrates around a few most likely paths joining $w_0$ and $w_f$ that are local maxima of the probability density functional. In other words, all probable paths can be obtained as a perturbation of a few critical paths (or activation trajectories). If the critical paths are sufficiently separated, we can estimate the total probability by approximating each cluster as a Gaussian centered around the cluster maximum.

The local maxima of $p(w(t)|w_0, t_0)$ can then be found by minimizing the action $S(w(t))$ in \cref{eq:msr} (or equivalently the simplified action in \cref{eq:decomposition}). Using the Euler-Lagrange equation $d/dt \partial_{\dot{\w}} \L(w,\dot{\w}) = \partial_\w \L(w,\dot{\w})$, we obtain that critical paths satisfy the differential equation
\begin{align*}
\ddot{\w}(t) = \nabla V(w(t)) = \nabla \bra{ \frac{1}{2} \nabla U(w(t))^2 - D \nabla^2 U(w(t)) },
\end{align*}
where the effective potential $V(w)$ is the same that appears in the decomposition in \cref{eq:decomposition}.
We observe that the Laplacian $ \nabla^2 U(w(t))$ of the potential $U(w)$ acts as a drag term in this expression. Therefore, depending on the temperature, the critical paths move more slowly when the curvature increases, which will play a role later.

For ease of exposition, let us assume for the moment that there is only one critical path $\w_c(t)$ between $w_0$ and $w_f$ that satisfies the above equations. Furthermore, let us assume that the path is along a coordinate axis, so that we can expand the potential up to second order around the path as
\[
U(u, \v) = a(u) + \frac{1}{2} \v \cdot b(u) \v.
\]
The Lagrangian associated with this process is
\begin{align}\label{eq:full_toy_Lagrangian}
\L(w, \dot{w})
=& \frac{1}{2D}\left\{ \frac{1}{2}[\dot{\v}^2 + b(u)\v]^2 - \tr[b(u)] \right\} + \frac{1}{2D}\left\{\frac{1}{2}[\dot{u}^2 + a'(u)]^2 - D a''(u) \right\} + \nonumber\\
&\left\{ \frac{[\v\cdot b'(u) \v]^2 }{16D} + \frac{[\dot{u}^2 + a'(u)][\v\cdot b'(u) \v]}{4D} - \frac{1}{2}\v\cdot b''(u) \v\right\}
\end{align}
The first term in eq.~\eqref{eq:full_toy_Lagrangian} accounts for the diffusion along the $\v$ direction. The third therm contains both derivatives of $b(u)$ and second-order terms in ($\v$); we can neglect it if we assume the validity of the saddle point approximation and that $b''(u) \ll b^2(u)/D$, that is, that the $b(u)$ varies slowly enough. Since we are mainly interested in the dynamics along the $u$ coordinate, we can integrate out the variable $\v$ from
eq.~\eqref{eq:msr}. We then obtain
\begin{equation}\label{eq:eq_marinalization_of_p}
p(u(t)|u_0, t_0) \approx  e^{-\frac{1}{2D}\int\left\{\frac{1}{2}[\dot{u}^2 + a'(u)]^2 - D a''(u) \right\}dt}\int_{\v(0)=0}^{\v(T)=0} e^{-\frac{1}{2D}\int\left\{ \frac{1}{2}[\dot{\v}^2 + b(u)\v]^2 - \tr[b(u)] \right\}dt} d\v
\end{equation}
When the diffusion in the $\v$ direction is much faster than the dynamics along $u$,
we can replace the integral with the local equilibrium distribution of $\v$ at a fixed $u$. The final expression for the marginalized probability density is
\begin{equation}\label{eq:eq_marginalization_of_p}
p(u(t)|u_0, t_0) \approx  e^{-\frac{1}{2}\log(2\pi|b|)}e^{-\frac{1}{2D}\int\left\{\frac{1}{2}[\dot{u}^2 + a'(u)]^2 - D a''(u) \right\}dt}.
\end{equation}

Under this approximation, and introducing an effective potential $U_\eff(w) = U(w) + D \log |\H U(w)|$ we finally obtain that the probability of reaching a point $w_f$ in a given time $t_f$ is given by
\begin{equation}
\label{eq:decomposition-marginalized}
p(\w_f,t_f|\w_0, t_0) = e^{-\frac{1}{2D} \Delta
U_\eff(w)}  \int_{\w_0}^{\w_f}  e^{-\frac{1}{2D}\int_{t_0}^{t_f} \frac{1}{2} \dot{u}(t)^2 + V(u(t)) dt} d u(t).
\end{equation}

This is critical, as it shows that both the speed and probability of convergence are controlled by the effective potential $U_\eff= U(w) - D \log |\H U (w)|$, which corrects the original potential by a term that depends on both the \emph{diffusion coefficient} (which scales as $D=k/B$, where $B$ is the batch-size and $k$ is a constant that depends on the architecture), and the \emph{curvature} (determinant of the Hessian) at that point.
That is, {\em to account for reachability, the potential needs to be corrected with the local curvature, and the amount of correction depends on the temperature.} One consequence of this is the often observed fact that sharp minima may not be minima at all for this particular potential when the temperature is sufficiently high (recall that, for a fixed learning rate, the diffusion coefficient scales as $D=k/B$, where $B$ is the batch-size and $k$ is a constant that depends on the architecture). Moreover, this suggests that the dynamic part of the potential can create spurious local minima that can inhibit learning of new problems in a transfer learning scenario \cite{achille2018critical}.

However, \cref{eq:decomposition-marginalized} still depends on the geometry of the optimization landscape, rather than properties intrinsic to the task.
We now connect the curvature to the amount of information needed to solve a task. Using this connection, we are able to characterize the {\em ``learnability''} (reachability)  of a task in terms of information-theoretic properties of the data. This completes our program of characterizing the geometry and topology of the space of tasks in a manner that, to first approximation, does not depend on how the task is actually learned.

\section{Information, Curvature and Kramer's Rate}
\label{sec:kramers}

To  establish a link between the curvature $U_\eff(w) = U(w) + D \log|\H U(w)|$ and the structure function of the task, note that when the network is trained with weight decay, with coefficient $\gamma$, the effective potential $U_\eff$ minimized by the network is given by:
\[
U_\eff = U + D \log |\H U (u)| = L_\D(w)  + \frac{\gamma}{2} \norm{w}^2 + D \log |\gamma I + H(w)|,
\]
where $H(w)$ is the hessian of the cross entropy loss $L_\D(w)$. By letting $\beta = 2\lambda^2 \gamma$ we obtain that the effective potential that affects the network while training with SGD is exactly the complexity $C_\beta(w;\D)$ of the dataset at level $\beta$. Therefore, we may rewrite the first (static) term of the transition probability in \cref{eq:decomposition-marginalized} as:
\begin{equation}
\label{eq:static-complexity}
p(\w_f,t_f|\w_0, t_0)_\text{static} = e^{-\frac{1}{2D} \Delta
C_\beta(\D;P,Q)}.
\end{equation}
This has the important implication that the transition probability is upper-bounded by a static part that depends solely on the complexity of the task, or more generally on the difference in complexity between tasks when fine-tuning. To this, however, we must add a dynamic term that also depends on the architecture of the network and the geometry of the loss landscape, and may in general be non-trivial and further reduce the reachability of a task.

From \cref{eq:decomposition-marginalized} and \Cref{eq:static-complexity}, we can derive the Kramer's convergence rate $1/\tau_K$, which is the expected time of convergence to a minimum, as
\begin{equation}
\label{eq:kramers-rate}
1/\tau_K = C e^{-\frac{1}{D}\Delta C_\beta(w;\D)}.
\end{equation}
That is, \emph{the expected time of convergence scales with the difference in complexities between tasks}.

\section{Empirical validation}
\label{sec:experiments}

\begin{figure}
    \centering
    \includegraphics[width=0.3\linewidth]{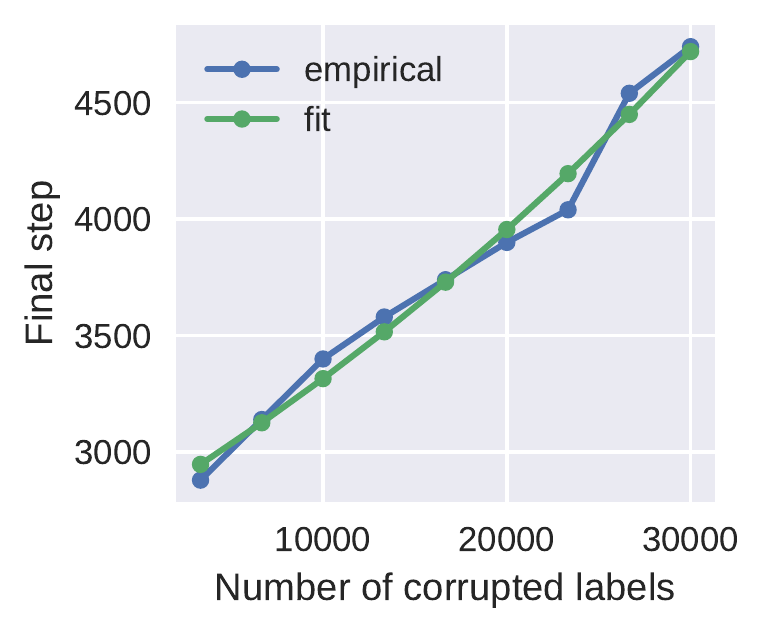}
    \includegraphics[width=0.3\linewidth]{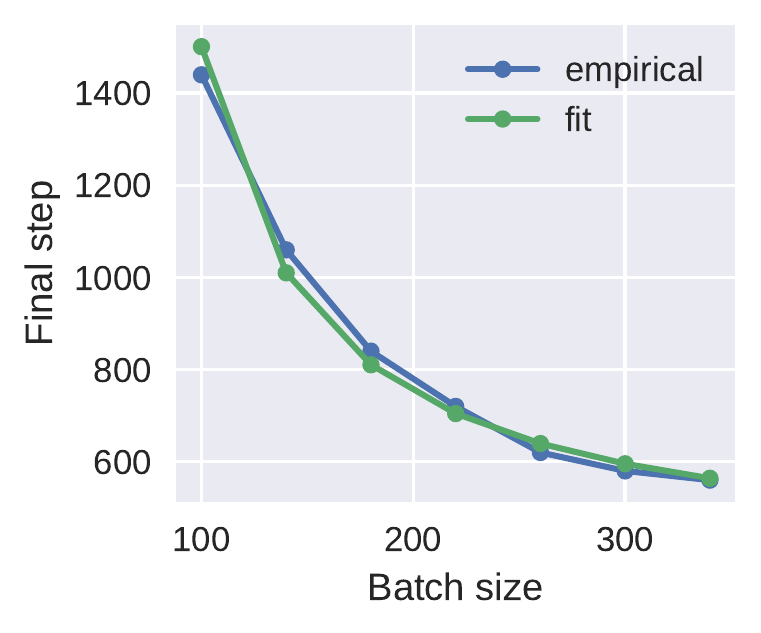}
    \includegraphics[width=0.34\linewidth]{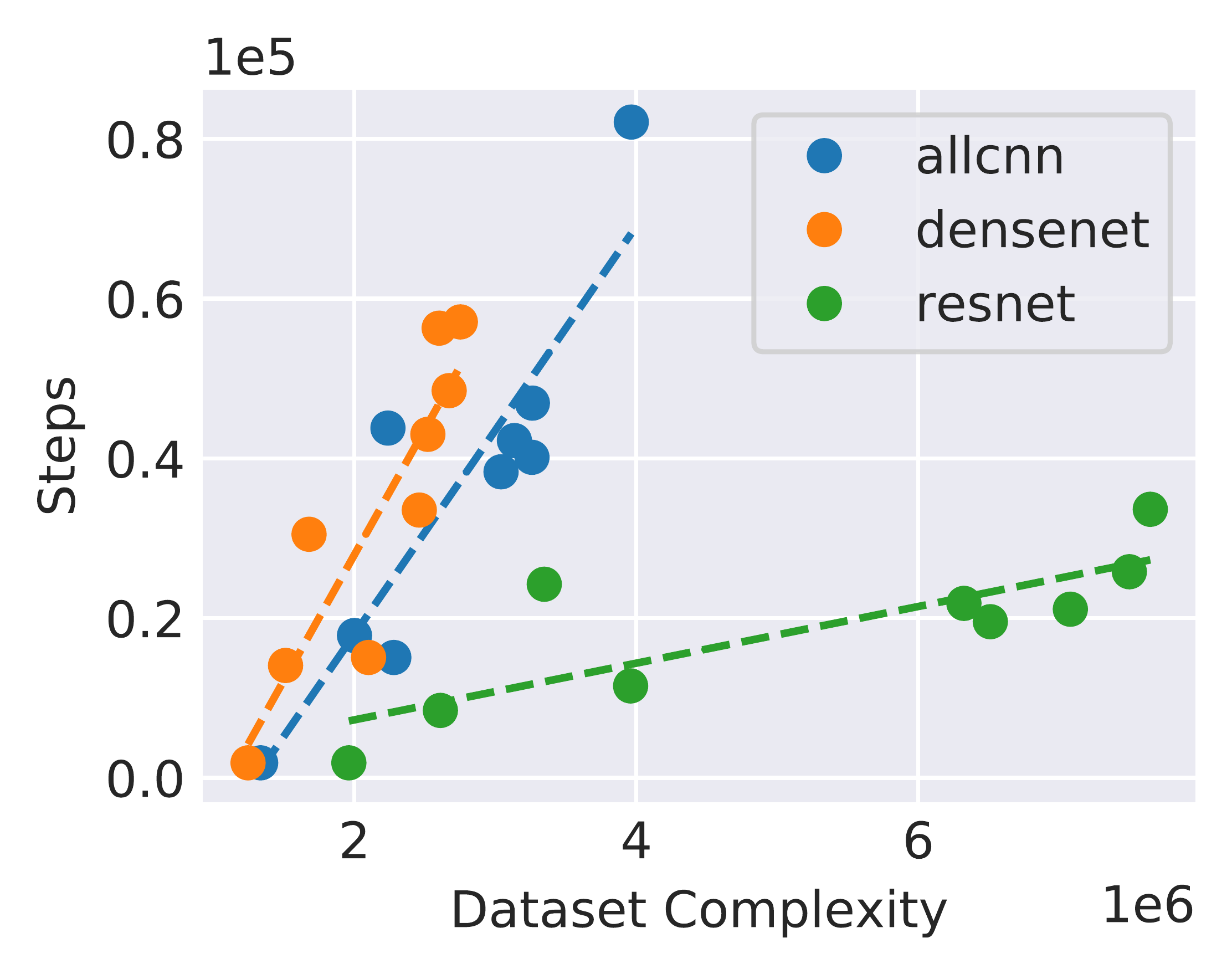}
    \caption{\textbf{(Left)} Plot of the time needed for an AlexNet model to converge, against the number of corrupted random labels added to CIFAR-10. Corrupting labels in the dataset increases the quantity of information that the network needs to store (memorize) in order to fit the dataset (minimize the loss below a given threshold.)
    Correspondingly, \cref{eq:eq_marginalization_of_p} predicts an increase of the time to convergence.
    Tthe trend of the empirical curve (blue) follows the theoretical prediction (green), where the parameters of the coefficients (\eg, dissipation constant $D$) are fitted from the data.
    \textbf{(Center)} Changing the batch size $B$ changes the dissipation constant $D \propto \eta/B$, and correspondingly, by \cref{eq:eq_marginalization_of_p}, changes the time to convergence.
    \textbf{(Right)} For several architectures, we plot the time needed for the network to converge (minimize the loss $U$ below a certain threshold) against the estimated complexity of the task, measured by \cref{eq:kl-complexity} using a Gaussian prior and posterior.
    }
    \label{fig:convergence-time}
\end{figure}

\subsection{Convergence time for different datataset}

In \Cref{sec:kramers} we have seen that the Kramer's rate for convergence, in first approximation and ignoring the contribution of the dynamic part of the transition probability, is given by \eqref{eq:kramers-rate}.
This gives an empirically verifiable law to test our model: In \Cref{fig:convergence-time} (right) we plot the time (number of SGD steps) needed by different architectures to converge on several different datasets. We can see that, as expected, different architectures have different parameters that regulate how the complexity affects the convergence time but for a fixed architecture and hyperparameters, the time to converge mainly depends on the complexity of the task alone.

\textbf{Random labels.} The case of random labels is of particular theoretical interest since, provided the value of $\beta$ is below a critical point to allow memorization of the label, the complexity $C_\beta(w;\D)$ scales linearly with the amount of random labels. In \Cref{fig:convergence-time} (left) we show that, in accordance with our model prediction, the time to converge scales with the complexity of the dataset, \textit{i.e.}, in this case with the amount of random labels in the dataset.

\textbf{Changing the batch size.} Another way we can act on the time to converge is to change the diffusion constant $D$ of the network: We know that for a fixed learning rate the diffusion constant scales as $D=c/B$, where $B$ is the batch size. \Cref{fig:convergence-time} (center) shows that changing the batch size changes the time to convergence, following the predicted trend.

\subsection{Time to fine-tune between tasks}

In the previous section we tested the relation between the complexity of the task and the time employed by the network to converge, starting from a random initialization. In practice, we may start from the minimizer of another task, rather than from a random initialization (fine-tuning). In this case, we expect the time to converge to depend not on the complexity of the task, but rather on the reachability of the new task from the previous task (\Cref{sec:reachability}).

\begin{figure}
\centering
\includegraphics[width=.28\linewidth]{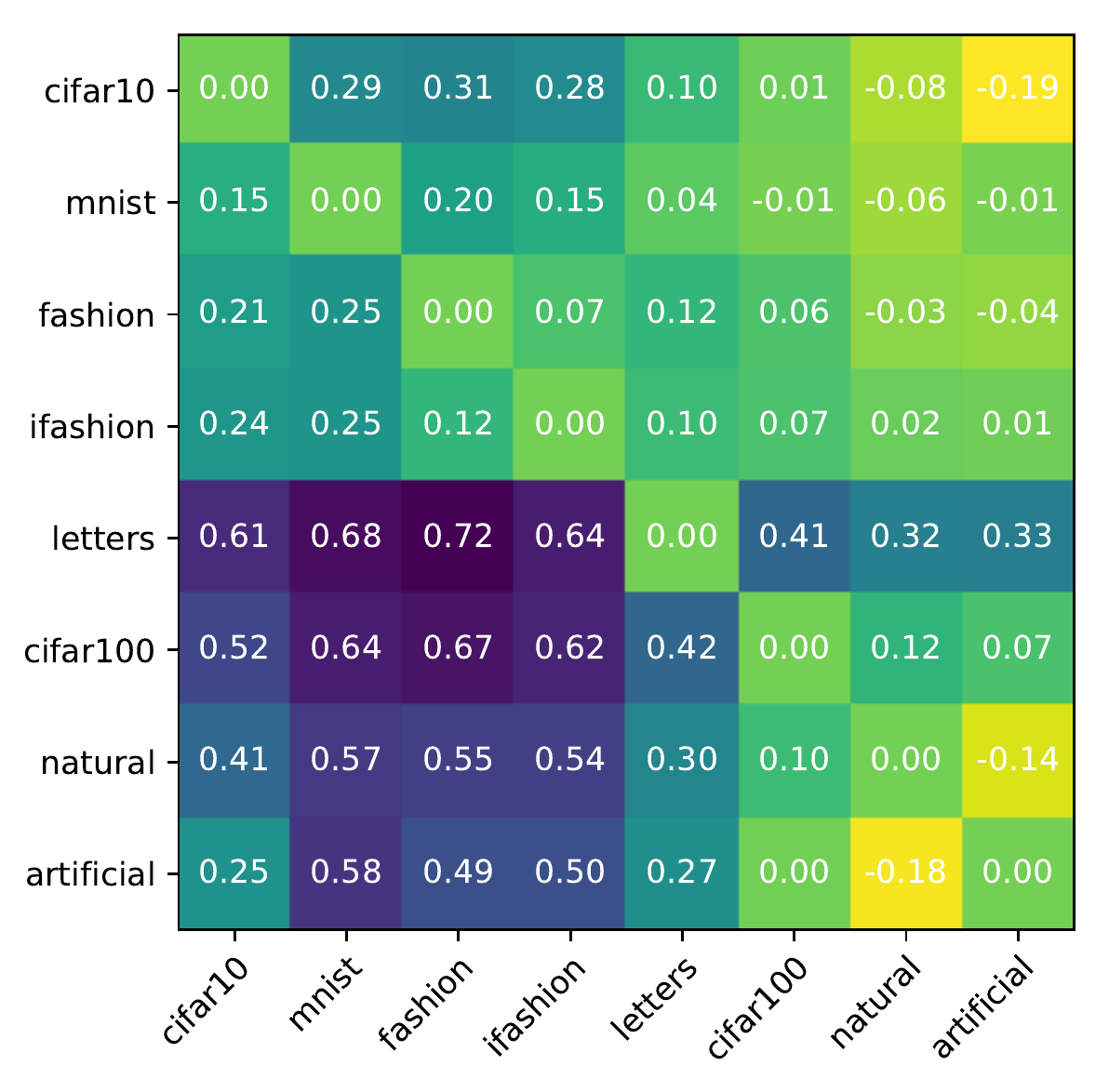}\
\includegraphics[width=.28\linewidth]{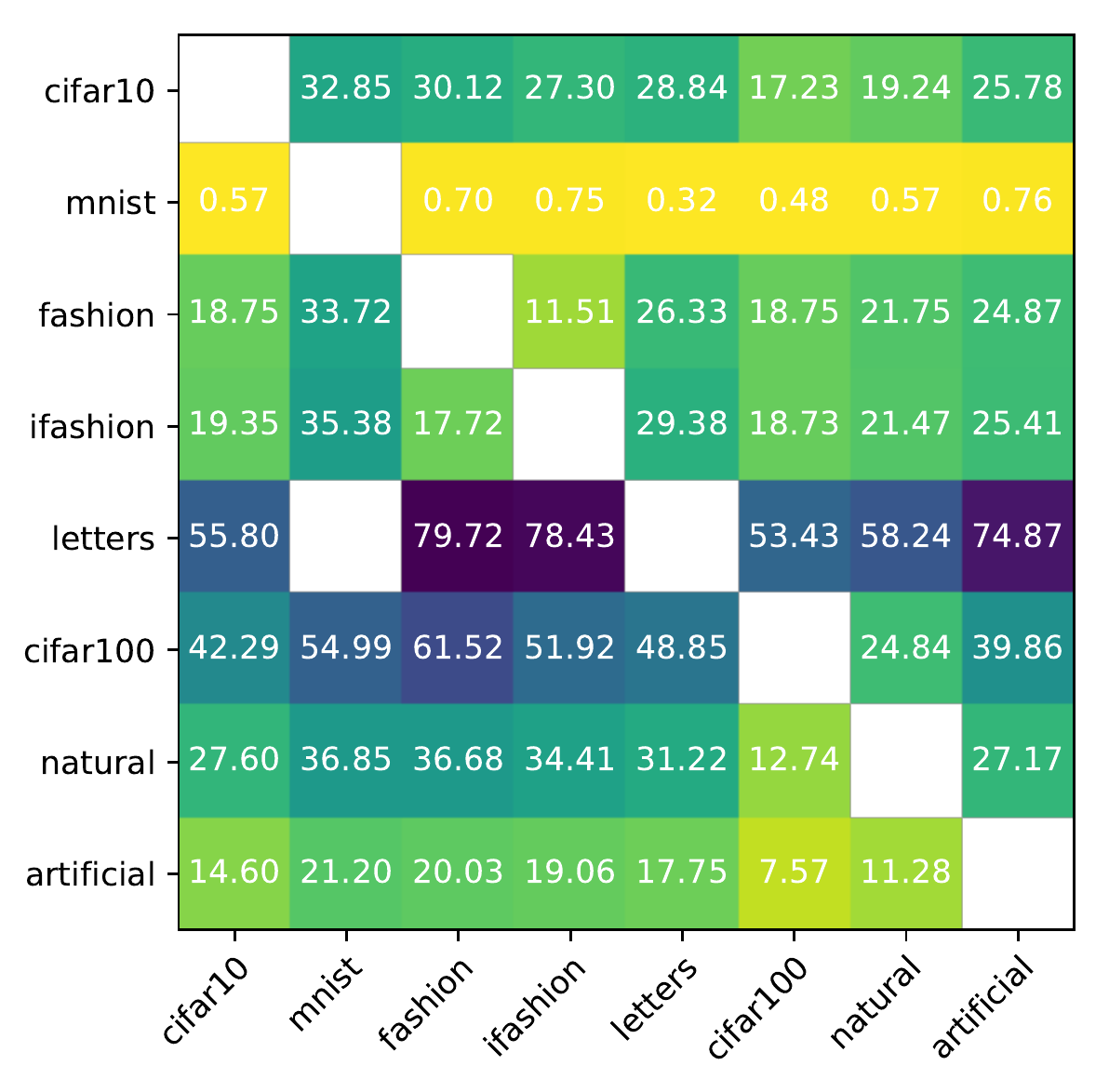}\
\includegraphics[width=.28\linewidth]{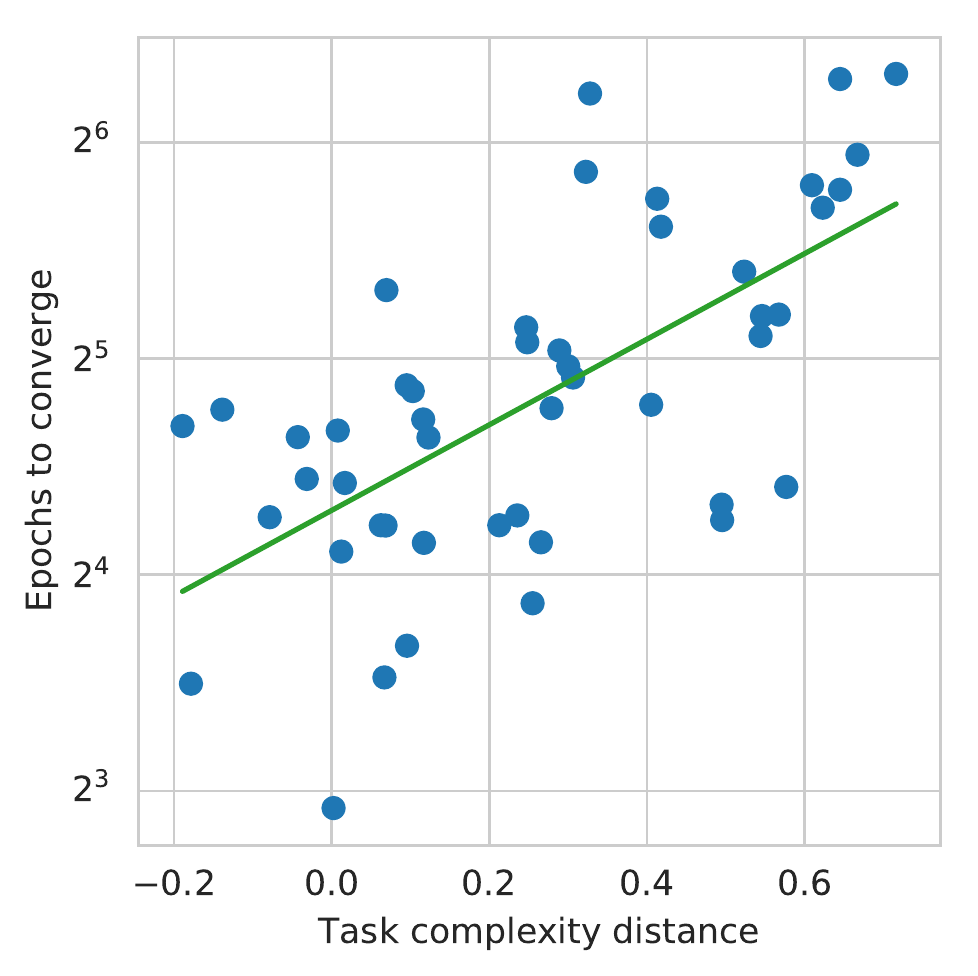}
\caption{
\textbf{(Left)} Reachability between tasks, based on the relative complexity. Each element of the matrix shows the time to convergence when fine-tuning from a pre-training classification task (columns) to a target task (rows). White cells denote no convergence. Notice that semantically similar task are close to each other, and that it is easier to go from a complex task to a related simple task than vice-versa. \textbf{(Center)} Training epochs necessary to fine-tune from one task (row) to another (column). \textbf{(Right)} Scatter plot of the relation between number of steps necessary to converge and the reachability of two datasets.
}
\label{fig:task-distance}
\end{figure}

In \Cref{fig:task-distance} (Left) we show for several popular datasets the reachability between tasks computed using the definition in \Cref{sec:reachability} and approximated with a ResNet-18 using \cref{eq:approximated-beta-complexity}. Notice that this matrix makes intuitive sense: semantically similar tasks are closer to each other, \textit{e.g.,}  CIFAR-100 is close to CIFAR-10 and to its two subsets of artificial and natural objects. Similarly, Fashion MNIST (fashion)  is  close to color inverted Fashion MNIST (ifashion) and to MNIST. Moreover the matrix captures the fact that it is generally easier to learn a task after training on a more complex, related, task (such as going from CIFAR-100 to CIFAR-10), rather than trying to learn a complex task starting from a simple one (\textit{e.g.,} going from MNIST to CIFAR-100).

From \cref{eq:kramers-rate} and \cref{eq:static-complexity} we know that the distance at level $\beta$ may be compared with the matrix of the time necessary to fine-tune from one task to another (\textit{i.e}., the training time until we reach some loss threshold), which we show in \Cref{fig:task-distance} (Center). In \Cref{fig:task-distance} (Right) we show the relation between time to fine-tune and reachability for several pairs of datasets, which again follows the theoretical prediction between the two.

\section{Discussion}

In this paper we have laid the foundations to enable quantifying the ease of transfer learning. This entails first defining and formally characterizing tasks, and then establishing some sort of topology in the space of tasks. To the best of our knowledge, we are the first to attempt this. We bring to bear tools from diverse fields, from Kolmogorov Complexity to quantum physics, to enable defining and computing sensible notions of distance that correlate with ease of transfer learning. In the process, we discover interesting connections between seemingly disparate concepts:
 The first is between the notion of task reachability, which we introduce, and the Kolmogorov Structure Function. This in turn is related to information-theoretic treatments of deep learning that have been recently developed \cite{achille2017emergence}. Furthermore, our analysis points to the importance of analyzing the dynamics of learning, rather than just focusing on the asymptotics, which confirms recent empirical discoveries in critical periods and the notion of Information Plasticity \cite{achille2018critical}.

\bibliographystyle{plain}
\bibliography{bibliography}

\end{document}